\documentclass{article}
\usepackage{spconf,amsmath,graphicx,enumitem,lipsum}

\title{DEEP MULTIMODAL SEMANTIC EMBEDDINGS FOR SPEECH AND IMAGES}
\name{David Harwath and James Glass}
\address{MIT Computer Science and Artificial Intelligence Laboratory\\
Cambridge, Massachusetts, 02139, U.S.A\\
\{dharwath, glass\}@mit.edu}

\begin{document}
\maketitle

\begin{abstract}
In this paper, we present a model which takes as input a corpus of images with relevant spoken captions and finds a correspondence between the two modalities. We employ a pair of convolutional neural networks to model visual objects and speech signals at the word level, and tie the networks together with an embedding and alignment model which learns a joint semantic space over both modalities. We evaluate our model using image search and annotation tasks on the Flickr8k dataset, which we augmented by collecting a corpus of 40,000 spoken captions using Amazon Mechanical Turk.
\end{abstract}
\begin{keywords}
Neural networks, multimodal semantic embeddings
\end{keywords}
\vspace{-3mm}
\section{Introduction and Related Work}
\label{sec:introduction}

Conventional automatic speech recognition (ASR) systems utilize training data in the form of speech audio with parallel text transcriptions. In this paper, we investigate what is possible to do if those text transcripts were replaced with relevant visual images. Given a dataset comprised of image scenes with accompanying spoken audio captions segmented at the word level, we propose a model capable of learning to associate spoken instances of the word "dog" with images of dogs, to name just one example. Our model relies on a pair of convolutional neural networks (CNNs), one for images and another for speech, along with an alignment and embedding model. The outputs of the networks provide fixed-dimensional representations of variable-sized visual objects and spoken words, which are then mapped into a shared semantic embedding space. This allows us to align the words in the captions to the objects they refer to in the image scene. While a large body of research on jointly modeling images and text exists in the literature, we are not aware of any prior work that models the semantics of images and speech directly on the audio signal level.

Multimodal modeling of images and text has been an extremely popular pursuit in the machine learning field during the past decade, with many approaches focusing on accurately annotating objects and regions within images. For example, Barnard et al. \cite{barnard_2003} relied on pre-segmented and labelled images to estimate joint distributions over words and objects, while Socher \cite{socher_2010} learned a latent meaning space covering images and words learned on non-parallel data. While these approaches focused on improving the identification of visual objects from a pool of predefined classes, other research has studied the problem of aligning text to the images or videos they describe. For example, Kong et al. \cite{kong_2014} took visual scenes with high level captions, parsed the text, detected visual objects, and then aligned the two modalities with a Markov random field. Lin et al \cite{lin_2014} aligned semantic graphs over text queries to relational graphs over objects in videos to perform natural language video search. Matuszek et al. \cite{matuszek_2012} employed separate classifiers over text and visual objects that shared the same label sets.

A related problem is that of natural language caption generation. While a large number of papers have been published on this subject, recent efforts using recurrent deep neural networks \cite{karpathy_2015, vinyals_2015} have made tremendous progress and generated much interest in the field. While our work in this paper does not aim to generate captions for images, it was originally inspired by the text-to-image alignment models presented by Karpathy in \cite{karpathy_2015, karpathy_2014}. In \cite{karpathy_2015}, Karpathy uses a refined version of the alignment model presented in \cite{karpathy_2014} to produce training exemplars for a caption-generating RNN language model that can be conditioned on visual features. Through the alignment process, a semantic embedding space containing both images and words is learned. Other works have also attempted to learn multimodal semantic embedding spaces, such as Frome et al. \cite{frome_2013} who trained separate deep neural networks for language modeling as well as visual object classification. They then embedded the object classes into a dense word vector space with the neural network language model, and fine-tuned the visual object network to predict the embedding vectors of the words corresponding to the object classes. This paper shares much of the same spirit as prior work on semantic embedding of images and words, but with a key difference - instead of dealing with text at the orthographic level, we learn a model which can derive meaning directly from spoken audio.

\section{Model Description}
\label{sec:models}
Our overarching goal is to be able to represent examples of spoken words, alongside examples of visual objects, as points in a single, high dimensional vector space. For example, in this vector space, we want different spoken examples of the word ``dog'' to neighbor one another, and also to neighbor image crops containing dogs. In order to do this, we require some means to transform variable sized image crops as well as variable duration audio waveforms into fixed dimensional vector representations. Further, we also require some way of coaxing these vectors into taking on the the property that semantically similar images and words neighbor one another. To achieve this, we employ two separate neural network architectures, one for images and one for audio, which we then marry together with an embedding alignment model.

\subsection{Region Convolutional Neural Network}
\label{ssec:rcnn}

In order to detect a set of candidate regions in an image which are likely to contain meaningful objects, we use the Region Convolutional Neural Network (RCNN) model \cite{girshick_2013}. The RCNN object detector works by first using selective search \cite{selective_search} to build a large list of proposal regions, typically numbering in the thousands for a given image. Each proposal region is then fed into a CNN object classifier, which is used to extract the activations of the penultimate layer of neurons in the network. These activations form a fixed-dimensional (4096 in \cite{girshick_2013}, as well as our work) feature vector representation of each proposal region. A set of one-versus-all support vector machines are then used to calculate detection scores over some set of classes for each region, and highly overlapping regions with similar classification scores are merged. Finally, the remaining set of regions can be ranked in order of their maximum classification score across all classes. In our work, we follow \cite{karpathy_2015} and take the top 19 detected regions along with the entire image frame, resulting in 20 regions per image. We use the $d_I = 4096$ dimensional RCNN feature vectors to represent each region, which we will refer to as $\mathcal{V} = \{v_i | i = 1 \dots 20\}$

\subsection{Spectrogram Convolutional Neural Network}
\label{ssec:scnn}
Previous efforts \cite{karpathy_2015, vinyals_2015} to perform semantic alignment of text to objects in image scenes have benefited from the fact that text is naturally segmented into words, and all instances of the same word share the same orthography. On the other hand, segmenting continuous speech into words is nontrivial, and different spoken instances of the same underlying word will inevitably differ in not only their duration, but also in their acoustic feature representations as influenced by factors such as the microphone and speaker characteristics and the context in which the word was spoken.

While a speech recognition system is a reasonable solution for building a spoken interface for natural language image retrieval systems such as the one described in \cite{karpathy_2015}, in this work we are more interested in investigating the potential of neural networks to learn meaningful semantic representations which operate directly on the feature level. However, tasking our system with also performing word segmentation on the audio stream significantly complicates the problem at hand. We choose to take a step back from the text-based framework by pre-segmenting each spoken caption into a sequence of audio waveforms, each containing a single ground-truth word, and then throwing away the word identity of each segment. 

In \cite{bengio_2014}, the authors trained a CNN isolated word recognizer and utilized it for N-best recognition hypothesis re-ranking; here, we propose to use a similar CNN to model the spectrogram of each isolated word in the image captions. Standard CNNs expect their inputs to be of a fixed size, so in order to accommodate our variable duration words we follow \cite{bengio_2014} and choose to embed their spectrograms in a fixed duration window, applying zero-padding and truncation when necessary. While \cite{bengio_2014} found that a 2 second window was sufficient to capture the duration of 97\% of the words in their corpus, in our case a 1 second long window is long enough to capture 99.9\% of the words appearing in our data. 

\begin{figure}[htb]
  \centering
  \centerline{\includegraphics[width=8.5cm]{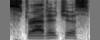}}
  \caption{Log mel filterbank spectrogram of the word ``strategists''}
  \label{fig:spectrogram}
\end{figure}

To create the spectrogram representing each word, we begin by performing forced-alignment of the audio to its ground truth text transcription in order to determine word boundary information. Next, we apply a standard 25 millisecond window with a 10 millisecond shift to each word utterance, extracting log energy filterbank features for each window using 40 filterbanks spaced along the mel scale. Next, we subtract the mean value and then apply variance normalization to the entire spectrogram. Finally, we either pad with zeros or truncate equally on both sides to force the spectrogram to have a width of 100 frames, or 1 second. Figure \ref{fig:spectrogram} shows an example of what the input data to the network looks like for an instance of the word ``strategists''.

We rely on the Caffe \cite{caffe} toolkit to train our networks and extract the word spectrogram features. Our CNN architecture is as follows:

\begin{enumerate}[itemsep=0mm]
\item Pixel-by-pixel mean image spectrogram subtraction, with the mean spectrogram estimated over the entire training set;
\item Convolutional layer with filters sized 5 frames by 40 features with a stride of 1, vertical padding of 1 pixel on both the top and bottom, and 64 output channels with a ReLU nonlinearity;
\item Local response normalization of width 5, $\alpha = 0.0001$, and $\beta = 0.75$;
\item Max pooling layer of height 3, width 4, vertical stride 1, and horizontal stride 2;
\item Two fully connected layers of 1024 units each, with a dropout ratio of 0.5 and ReLU nonlinearities;
\item A softmax classification layer
\end{enumerate}

\begin{figure}[htb]
  \centering
  \centerline{\includegraphics[width=8.5cm]{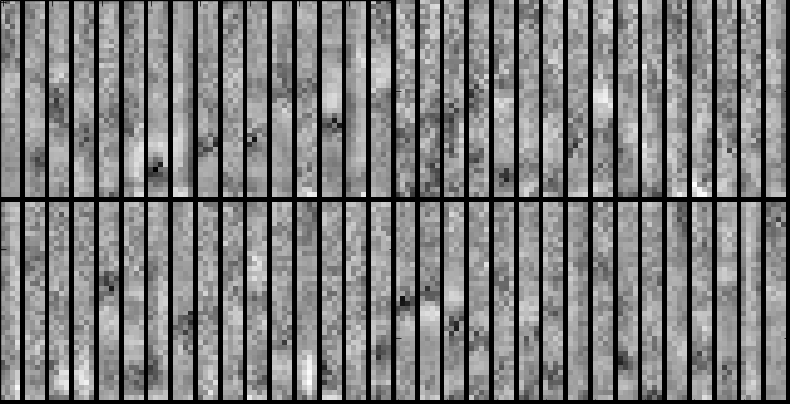}}
  \caption{64 learned filters for the spectrogram CNN}
  \label{fig:cnn_filters}
\end{figure}

To extract vector representations for each word in some image caption, we feed the word's spectrogram through the network and discard the softmax outputs, retaining only the activations of the $d_W = 1024$ dimensional fully connected layer immediately before the classification layer. For a given caption, we will refer to these vectors as $\mathcal{W} = \{w_j | j \dots N_w \}$, where $N_w$ is the number of words appearing in the caption.

\subsection{Embedding Alignment Model}
\label{ssec:alignment}

Given an image-caption pair and their corresponding object detection boxes and word spectrograms, our task is to align each word with one of the detection boxes found in the image. To do this, we adopt the transform model from \cite{karpathy_2014} but with the objective function presented by \cite{karpathy_2015}. However, we replace the text modelling side of Karpathy's models with our word spectrogram CNN, enabling us to align the image fragments directly to segments of speech audio. We provide a brief overview of the alignment model and objective here.

Let $\mathcal{V} = \{v_i | i = 1 \dots 20\}$ be the set of $d_I$-dimensional vectors representing the activations of the penultimate layer of the RCNN for each detected image region, as described in Section \ref{ssec:rcnn}. Also let $\mathcal{W} = \{w_j | j \dots N_w\}$ be the $d_W$-dimensional vectors representing the similar activations of the spectrogram CNN on each of the $N_w$ words in the spoken caption. The job of the alignment model is to map all of the $v \in \mathcal{V}$ and $w \in \mathcal{W}$ vectors into a shared, $h$-dimensional space where semantically related words and images have a high similarity.

The alignment model is two-faceted, with separate transforms applied to the image vectors as well as the word spectrogram vectors. We use an affine transform, $y = W_{m} v + b_{m}$ to map an image vector $v$ into the $h$-dimensional semantic embedding space. To map a word spectrogram vector $w$ into that same embedding space, we use a nonlinear transform, $x = f(W_{d} w + b_{d})$ where $f(z)$ is some element-wise nonlinear function. For the experiments in this paper, we set $f(z) = \max(0, z)$.

Motivated by the assumption that the spoken caption $l$ for a given image $k$ should contain words which directly reference objects in the image, Karpathy's objective function tries to assign a high similarity to matching image-caption pairs by ``grounding'' each word vector to one or more image fragment vectors. The inner product similarity between a given word embedding and an image fragment embedding is used to measure the degree of grounding, and each word in caption $l$ is given a score according to its maximum similarity across all image fragments from image $k$. An overall image-caption similarity score is then computed by summing the scores of all words in the caption, thresholded below at 0:
\begin{equation}
S_{kl} = \sum_{t \in g_l}{\max_{i \in g_k}(0, y_{i}^{T}x_t)},
\end{equation}
where $g_l$ denotes the set of image fragments in image $l$, and $g_k$ is the set of word spectrograms in caption $k$.

In \cite{karpathy_2015}, Karpathy uses a max margin objective function which forces matching image-caption pairs to have a higher similarity score than mismatched pairs, by a margin. Given that $S_{kk}$ denotes the similarity between a matching image-sentence pair, the cost is defined as:
\begin{equation}
\begin{split}
\mathcal{C}(\theta) = & \sum_{k}\Big[ \sum_{l}{\max(0, S_{kl} - S_{kk} + 1)} \\
& + \sum_{l}{\max(0, S_{lk} - S_{kk} + 1)} \Big].
\end{split}
\end{equation}
In practice, we use stochastic gradient descent to optimize this cost function in terms of the parameters $\theta = \{W_m, b_m, W_d, b_d\}$.

\vspace{-3mm}

\section{Data}
\label{sec:data}

Recent works on natural language image caption generation \cite{karpathy_2015, vinyals_2015} have utilized a number of datasets which contain images alongside human-generated text captions, such as Pascal, Flickr8k \cite{flickr8k}, Flickr30k \cite{flickr30k}, and MSCOCO \cite{mscoco}. However, none of these datasets include any speech data, so we decided to collect our own spoken audio for our experiments. Because of its manageable size and ubiquitousness in the previous literature, we choose to use the Flickr8k as the starting point for our data collection. 

Flickr8k contains approximately 8,000 images captured from the Flickr photo sharing website, each of which depicts actions involving people or animals. Each image was annotated with a text caption by five different people, resulting in a total of 40,000 captions. To collect these captions the authors turned to Amazon's Mechanical Turk, an online service which allows requesters to post "Human Intelligence Tasks" (HITs). These HITs are then made available to anonymous, non-expert workers, or "Turkers", who can choose to complete the tasks for a small amount of money. We turned to Mechanical Turk to collect spoken audio recordings for each of the 40,000 captions from the Flickr8k dataset. We use the Spoke JavaScript framework \cite{saylor_2015} as the basis of our audio collection HIT. Spoke is a flexible framework for creating speech-enabled websites, acting as a wrapper around the HTML5 getUserMedia API while also supporting streaming audio from the client to a backend server via the Socket.io library. The Spoke client-side framework also includes an interface to Google's SpeechRecognition service, which can be used to provide near-instantaneous feedback to the Turker. 

\begin{figure}[htb]
  \centering
  \centerline{\includegraphics[width=8cm]{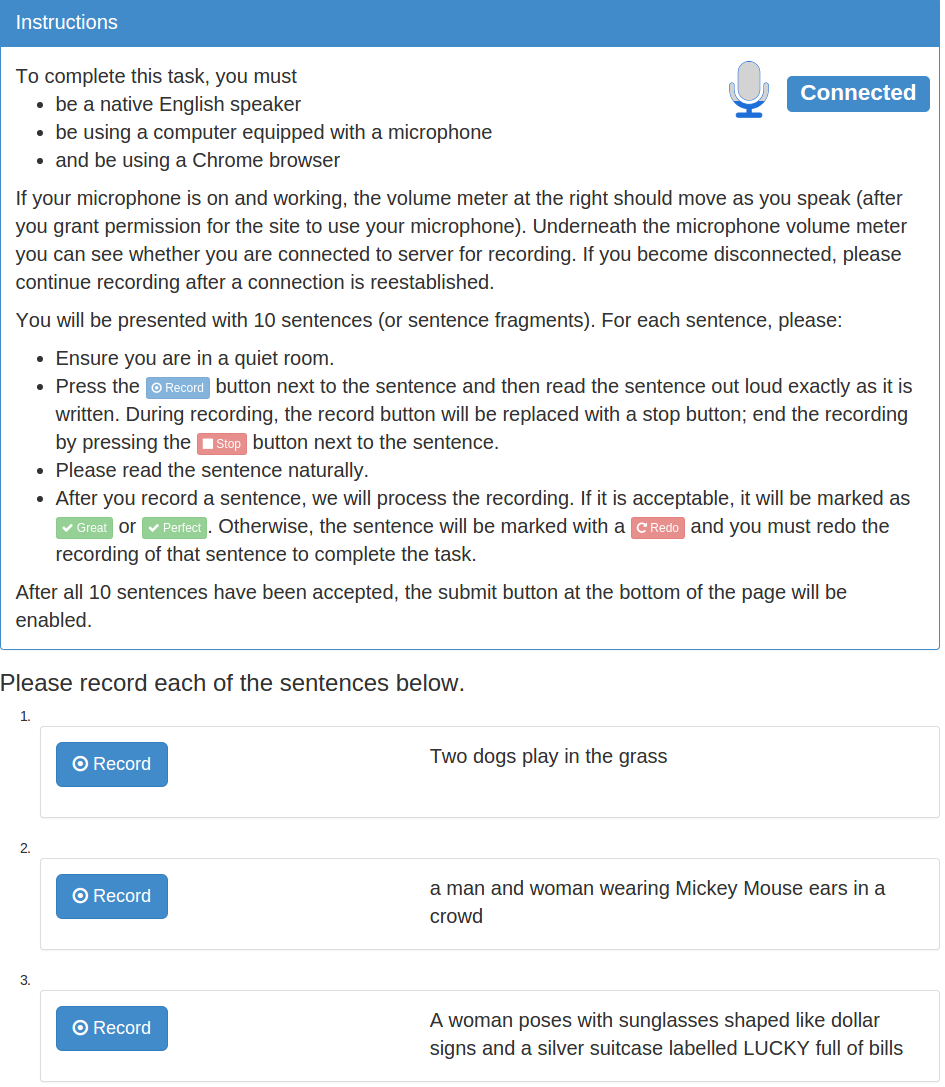}}
  \caption{Audio collection interface for capturing spoken captions on Amazon Mechanical Turk.}
  \label{fig:spoke_interface}
\end{figure}

Figure \ref{fig:spoke_interface} displays a screenshot of the audio collection interface we used in our HITs. A set of 10 random captions are displayed to the user, who can click the start/stop button to record their speech while they read each caption out loud. A playback button allows the Turker to listen to their own recordings and diagnose any problems with their microphone or environment. Spoke pipes the audio to the Google recognizer, checks the recognition result against the prompt, and notifies the user if their speech could not be recognized accurately. The Turker is then given the option to re-record the errorful caption. The HIT cannot be submitted until all 10 captions have been successfully recorded. In our experiments, we use a very simple metric for verification - 60\% or more of the caption words must appear in the recognition result, regardless of ordering. We found this to be both lenient and sufficient - users rarely complained about the system correctly recognizing their speech, and 95.7\% of the collected utterances were easily aligned to their caption text using our Kaldi \cite{kaldi} forced alignment system. The majority of the utterances flagged as unalignable were either empty or cut short, which we believe may have been due to client-server connection issues; the problematic utterances were recollected by another round of HITs. We paid the turkers 0.5 cents per spoken caption, resulting in at total cost of just over \$200 including Amazon's service fee. We collected speech from 183 unique Turkers, with the average worker completing 218 captions. There were a handful of Turkers who completed far more than the average number of captions, with the highest number collected from a single worker being 2,978.

To further verify the integrity of our collected audio data, we split the 40,000 utterances into a 30,000 utterance training set, a 5,000 utterance development set, and a 5,000 utterance testing set, covering a 8,918 word vocabulary. Our splits correspond with the training, validation, and testing splits given by \cite{flickr8k}. We then used Kaldi to build a large vocabulary speech recognition system, adapting the standard Wall Street Journal recipe for a GMM/HMM + 
LDA + MLLT + SAT system for our data. We employed the training set to train the acoustic and language models, the CMU pronunciation lexicon, and the development set to tune the acoustic and language model weights. The final word error rate of our system on the test set was 11.67\%, providing another indication that our data is relatively high quality. In order to preprocess the Flickr8k data for our CNN, we employ this recognizer to force align the audio to the ground truth text transcripts and segment the audio at the word level. 

Because the Flickr8k corpus contains a small number of images and captions relative to datasets such as ImageNet \cite{imagenet}, we follow the example of \cite{karpathy_2015} and use the off-the-shelf RCNN provided by \cite{girshick_2013} trained on ImageNet to extract the 4096-dimensional visual object embeddings. Similarly, we employ supervised pretraining for the word spectrogram CNN using the Wall Street Journal SI-284 split \cite{wsj}. This set contains roughly 82 hours of speech, from which we extracted all instances of words occuring at least 10 times in the data. This gave us a total of 612,108 words covering a vocabulary of size 6,010, which we split 80/20 into training and testing sets. We used this data to train our word spectrogram CNN using the 6,010 word vocabulary as our output targets. Even though this training is supervised, 6,749 of the unique words appearing in the Flickr8k transcriptions (75\% of the vocabulary) do not appear in the training set for the spectrogram CNN.

\section{Experiments}
\label{sec:experiments}
We use stochastic gradient descent with a learning rate of 1e-6 and momentum of 0.9 across batches of 40 images to train our embedding and alignment model, and run our training for 20 epochs. Training is performed using the standard 6,000 image train set from the Flickr8k data, using the accompanying 30,000 captions. At each batch, we randomly choose to use only one of the five captions associated with each image. We tried several different settings for $h$, the dimension of the semantic embedding space, and found that values between 512 and 1024 seemed to work well, in line with \cite{karpathy_2014}. We also found that it was necessary to normalize the $w$ vectors to unit magnitude in order to prevent exploding gradients.

\begin{figure*}
    \begin{minipage}[b]{1.0\linewidth}
    \centering
      \begin{tabular}{cc}
        \includegraphics[width=8.5cm]{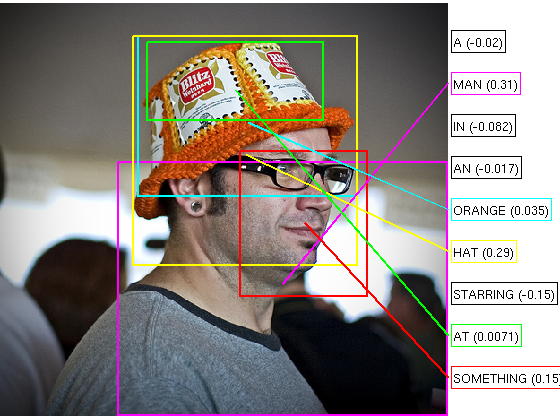} & \includegraphics[width=8.5cm]{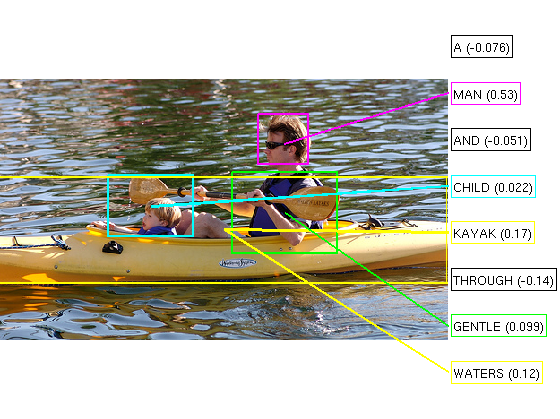} \\         \includegraphics[width=8.5cm]{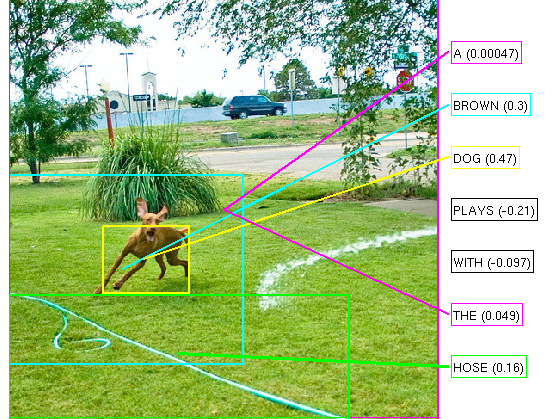} & \includegraphics[width=8.5cm]{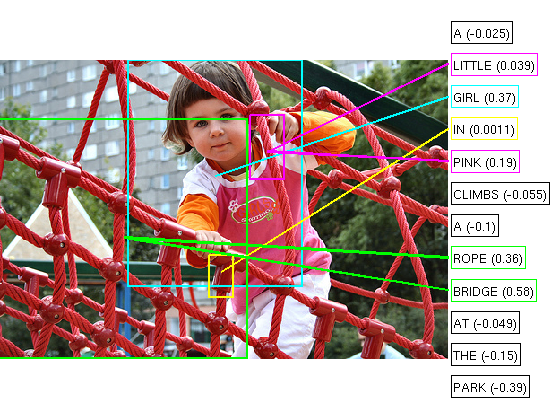} \\
      \includegraphics[width=8.5cm]{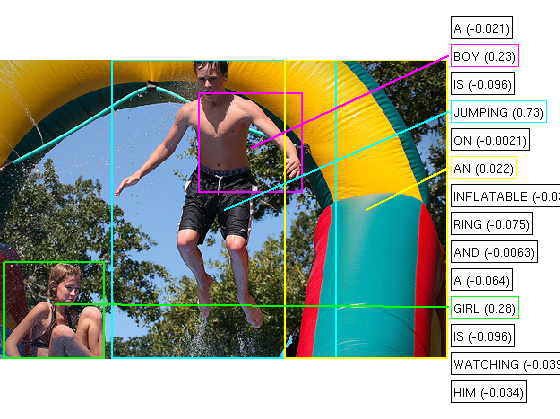} & \includegraphics[width=8.5cm]{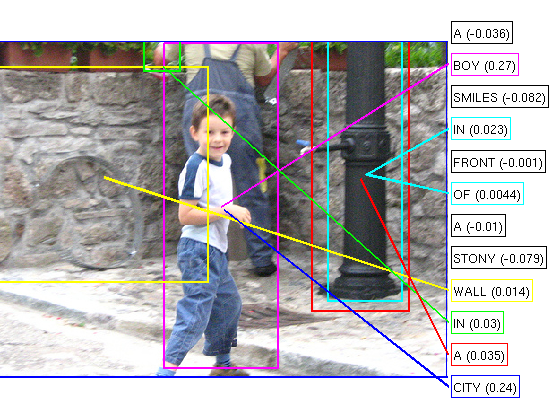}
      \end{tabular}
      \caption{Some examples of inferred alignments on the Flickr8k data. The words for each image's caption are stacked to the right of each image, accompanied by their alignment scores. To keep the images free from too much clutter, we threshold the scores at 0, displaying a link between the word and its maximally associated object bounding box only when its score is positive. Note that the system does not actually see the text of the caption words - only a spectrogram. We replace the spectrogram in these figures with the ground truth text for the sake of clarity.}\label{fig:visualizations}
    \end{minipage}%
\end{figure*}

To evaluate the alignment and embedding model, we follow the example of \cite{karpathy_2015, karpathy_2014, socher_2014} and use our model to perform image retrieval and annotation. Image search is defined as choosing a caption from the test set and then asking the system to find which image belongs with the caption. Image annotation is the opposite problem: choosing an image from the test set without its caption, and then asking the system to search over all the captions in the test set and find one of the five which belongs with the image. We report recall@10 as our evaluation metric, or the probability that the correct result is found in the top 10 returned hits. Table \ref{tab:results} details the results of our system (``Spectrogram CNN''), as well as a comparison to replacing the word spectrogram embeddings with 200-dimensional word vectors taken from \cite{huang_2012}. We also compare to Socher et al. \cite{socher_2014} and Karpathy \cite{karpathy_2014}. While our text + word vector system outperforms \cite{karpathy_2014}, the model is more similar to Karpathy's refinements made in \cite{karpathy_2015} but with a single layer word embedding network rather than a bidirectional recurrent neural network. \cite{karpathy_2015} reports high recalls on the Flickr30k data (50.5 search and 61.4 annotation), but does not include any results on the Flickr8k data. Although our spectrogram CNN does not perform nearly as well as any of the systems with access to the ground truth text, it massively outperforms a random ranking scheme. This is in spite of the fact that not only does the spectrogram CNN system not have direct access to the ground truth word identity of the caption words, but also that the CNN word embedding vectors are of dimension 1024 rather than 200.  We believe that these results are quite promising, and with more training data we expect to see substantial improvements. Figure \ref{fig:visualizations} displays several alignments of Flickr8k images to their captions inferred by our system. While by no means perfect, our system reliably aligns salient objects in the images with their associated caption words. 

We also trained several different word spectrogram CNNs with varying configurations. Table \ref{tab:cnn_accuracies} displays the top-1 and top-5 accuracies of a few of these networks. A two-layer conventional DNN with 1024 units per layer and ReLU nonlinearities achieved a classification accuracy of 75.5\%, while adding a third layer brought that number even lower to 69.5\%. We speculate that our training set is not large enough to train such a network. However, replacing the first fully connected layer with a 64-unit convolutional layer (following the architecture described in Section \ref{ssec:scnn}) boosted the accuracy to 84.2\%. We also trained a network with two convolutional layers and one fully connected layer and achieved similar results to the network with only a single convolutional layer. We also explored varying the size and shapes of the convolutional filters, pooling layers, and dimension of the fully connected layers, but the network achieving 84.2\% accuracy reflects our best performance. Although these networks show a wide range of top-1 accuracies, it is interesting to note that their top-5 accuracies are all in excess of 90\%. Figure \ref{fig:cnn_filters} displays the 64 filter responses from the first layer of our network.

\begin{center}
\begin{table}
 \begin{tabular}{||c c c||} 
 \hline
 Model & Search R@10 & Annotation R@10\\
 \hline\hline
 Socher et al. \cite{socher_2014} & 28.6 & 29.0 \\
 \hline
 Karpathy \cite{karpathy_2014} & 42.5 & 44.0 \\
 \hline
 Text + word vec & 49.0 & 56.7 \\ 
 \hline
 Spectrogram CNN & 17.9 & 24.3\\
 \hline
\end{tabular}
\caption{Image search and annotation results on the Flickr8k test images (1000 images with 5 captions each).}
\label{tab:results}
\end{table}
\end{center}

\begin{center}
\begin{table}
 \begin{tabular}{||c c c||} 
 \hline
 Model & Top-1 Acc. & Top-5 Acc. \\
 \hline\hline
 DNN, 2x1024 FC & 75.5 & 93.9 \\ 
 \hline
 DNN, 3x1024 FC & 69.5 & 91.4 \\
 \hline
 CNN, 1x64 Conv + 2x1024 FC & 84.2 & 97.4 \\ 
 \hline
\end{tabular}
\caption{Isolated word recognition accuracies on our WSJ test set. ``FC'' stands for ``fully connected''.}
\label{tab:cnn_accuracies}
\end{table}
\end{center}

\vspace{-2cm}
\section{Conclusion}
\label{sec:conclusion}

In this paper, we have presented our first efforts to construct a model which can learn a joint semantic representation over spoken words as well as visual objects. At training time, the model only requires weak labels in the form of paired images and natural language spoken captions. Our system aligns salient visual objects in the images with their associated caption words, in the process building a semantic representation across both modalities. We evaluate our model on the Flickr8k image search and annotation tasks, and compare it to several systems with access to the ground truth text.

There are many avenues which we would like to take this research next. Deeper investigation of the performance gap between CNN speech embedding and ground truth text systems is a logical first step, and increasing the amount of training data may shed some light on this. We would also like to incorporate word level segmentation into the alignment scheme, alleviating the need to use forced alignment and making our setting more realistic. Lastly, while the neural networks used to extract features for both the visual objects and the spoken words are pre-trained in a supervised fashion on outside data, we believe that with a very large amount of data it may be possible to train them together along with the alignment and embedding model.

\clearpage

\end{document}